\documentclass[11pt]{article}

\usepackage[final]{acl}

\usepackage{times}
\usepackage{latexsym}

\usepackage[T1]{fontenc}

\usepackage[utf8]{inputenc}

\usepackage{microtype}

\usepackage{inconsolata}

\usepackage{graphicx}

\usepackage{amsmath}

\usepackage{algorithm}
\usepackage{algorithmic}

\usepackage{booktabs}
\usepackage{multirow}
\usepackage[normalem]{ulem}
\useunder{\uline}{\ul}{}
\usepackage{arydshln}

\title{FroM: Frobenius Norm-Based Data-Free Adaptive Model Merging}

\author{
  Zijian Li$^{\dag}$ \quad Xiaocheng Feng$^{\dag \ddag}$\thanks{Corresponding authors.} \quad Huixin Liu$^{\dag}$ \\
  {\bf Yichong Huang$^{\dag}$} \quad {\bf Ting Liu$^{\dag}$} \quad {\bf Bing Qin$^{\dag \ddag}$\footnotemark[1]} \\
  $^{\dag}$Harbin Institute of Technology \quad
  $^{\ddag}$Peng Cheng Laboratory \\
  \texttt{\{lizijian, xcfeng, hxliu, ychuang, tliu, qinb\}@ir.hit.edu.cn}
}

\begin{document}
\maketitle
\begin{abstract}
With the development of large language models, fine-tuning has emerged as an effective method to enhance performance in specific scenarios by injecting domain-specific knowledge.
In this context, model merging techniques provide a solution for fusing knowledge from multiple fine-tuning models by combining their parameters.
However, traditional methods often encounter task interference when merging full fine-tuning models, and this problem becomes even more evident in parameter-efficient fine-tuning scenarios.
In this paper, we introduce an improvement to the RegMean method, which indirectly leverages the training data to approximate the outputs of the linear layers before and after merging.
We propose an adaptive merging method called FroM, which directly measures the model parameters using the Frobenius norm, without any training data.
By introducing an additional hyperparameter for control, FroM outperforms baseline methods across various fine-tuning scenarios, alleviating the task interference problem.
\end{abstract}

\section{Introduction}

In recent years, significant advancements have been made in the fields of artificial intelligence and natural language processing, with large language models (LLMs) emerging as a focal point of attention \cite{achiam2023gpt, guo2025deepseek}.
Among the various approaches, the pretraining and fine-tuning paradigm has emerged as the foundational framework for LLM technology \cite{zhao2023survey, han2021pre, parthasarathy2024ultimate}.
With a growing number of fine-tuned downstream models, model merging has emerged as an effective approach to integrate models trained under different hyperparameter settings or for multi-task learning \cite{jin2022dataless}.
Moreover, model merging is commonly implemented in a data-free method, making it particularly advantageous for privacy-sensitive applications like federated learning \cite{mcmahan2017communication, wang2020federated}.
\begin{figure}[!tbp]
  \centering
  \includegraphics[width=0.95\columnwidth]{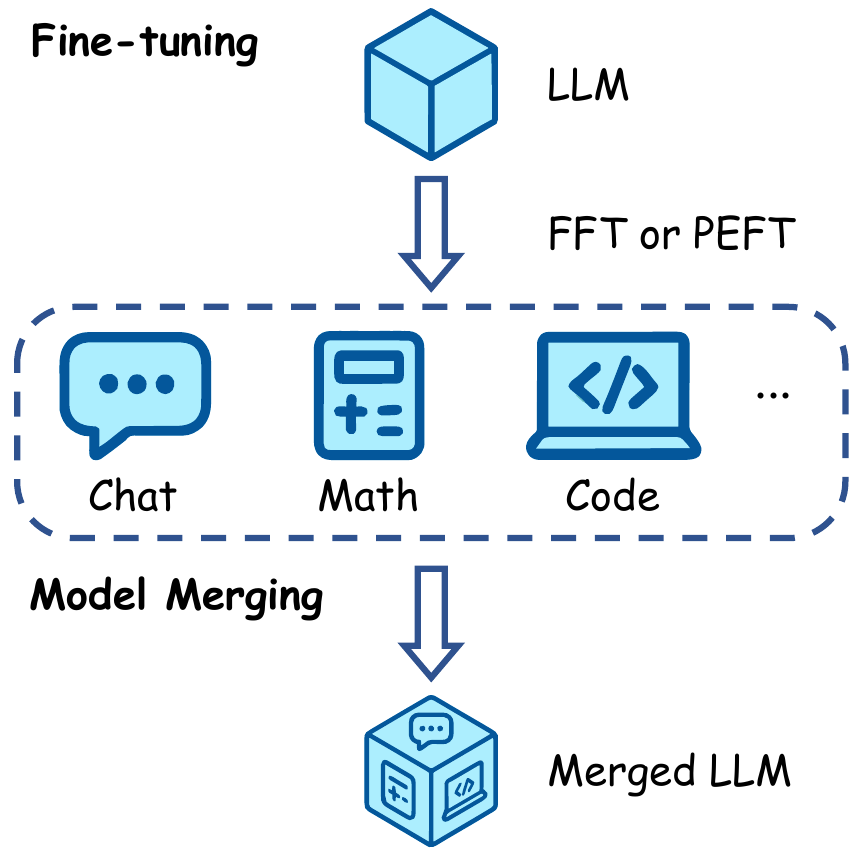}
  \caption{Illustrative diagrams of fine-tuning and model merging.}
  \label{fig:schematic}
\end{figure}

Most existing model merging methods are based on linear operations such as weighted averaging \cite{wortsman2022model} and task arithmetic \cite{ilharco2022editing}.
Weighted averaging combines parameters from multiple fine-tuned models, while task arithmetic enables multi-task learning and forgetting by adding and subtracting task vectors derived from parameter differences.
Nevertheless, these methods encounter a fundamental challenge: simple linear combinations of parameters often lead to task interference \cite{yadav2024ties}.
This issue is especially evident in parameter-efficient fine-tuning (PEFT) scenarios \cite{ding2023parameter}, such as Low-Rank Adaptation (LoRA) \cite{hu2021lora}.
Compared to full fine-tuning (FFT), the sparse structure of low-rank matrices in these methods further aggravates task interference \cite{tang2023parameter, stoica2024model}.
Figure~\ref{fig:schematic} presents a schematic illustration of the fine-tuning and model merging processes.

To address this challenge, we take inspiration from the nonlinear weighted optimization method of RegMean \cite{jin2022dataless} and introduce a novel method, \textbf{Fro}benius Norm-Based Data-Free Adaptive Model \textbf{M}erging (FroM).
Our method improves upon the RegMean method, which derives a closed-form solution by approximating the linear layer outputs of the models using the inner product matrix of the training data.
However, RegMean still encounters challenges due to its reliance on access to training data, which may not always be available.
To address this limitation, FroM adopts a data-free approach that directly measures model parameters using the Frobenius norm, and employs adaptive coefficients to adjust the discrepancies between task vectors.
Our approach reduces the reliance on training data, while outperforming baselines in both full fine-tuning and LoRA scenarios, effectively mitigating the task interference issue.

Our main contributions are as follows:
\begin{enumerate}
\item We propose an adaptive model merging method, FroM, which utilizes the Frobenius norm to measure the differences between models to be merged, without any training data.
\item Compared to baseline methods, our approach is broadly applicable to FFT and PEFT models, such as LoRA, and achieves optimal or suboptimal results in various scenarios.
\end{enumerate}

\section{Related Work}

\paragraph{Fine-tuning of LLMs.}
Fine-tuning is a method for optimizing pre-trained LLMs for specific tasks.
Through fine-tuning, LLMs not only excel in general tasks but can also meet the requirements of particular application scenarios \cite{wei2021finetuned, sanh2021multitask, chung2024scaling}.
From the perspective of the amount of parameters updated, fine-tuning can be categorized into two approaches: FFT and PEFT.
FFT updates all parameters of the pre-trained model, which often yields excellent performance on specific tasks.
However, this approach is both time-consuming and computationally intensive.
As models become larger and more complex, conducting FFT has become more challenging, particularly in environments with limited computational resources \cite{lv2023full}.
Alternatively, PEFT focuses on updating a small number of parameters to reduce computational costs and storage requirements, making it more widely used in resource-constrained settings or customized scenarios \cite{xu2023parameter, han2024parameter}.
One of the most widely used methods within PEFT is LoRA \cite{hu2021lora}, which significantly reduces computational and storage costs by fine-tuning low-rank matrices, achieving highly effective fine-tuning with a limited number of parameters.

\paragraph{Model Merging.}

\cite{lu2024merge} introduce several collaboration strategies for LLMs, highlighting model merging as a technique that prevents models from getting trapped in local optima and improves multi-task collaboration ability.
Traditional methods typically rely on weighted averaging of model parameters.
\cite{wortsman2022model} introduce Model Soups, which apply both a weighted averaging method and a greedy merging strategy.
\cite{matena2022merging} adopt a different approach by utilizing the Fisher Information Matrix for parameter-level model merging, applying a weighted merging strategy to enhance model integration.
Focusing on the outputs of linear layers, \cite{jin2022dataless} propose RegMean, a method that indirectly utilizes the training data through the inner product matrices to derive a closed-form solution for merging model parameters.
Meanwhile, \cite{ilharco2022editing} present a novel paradigm called task arithmetic, in which task vectors capture the ability shift of different fine-tuned models compared to the pre-trained model across various tasks.
This allows for operations like forgetting and learning through arithmetic manipulation of the task vectors.

However, the task vectors of different models may inevitably lead to some degree of task interference.
Prior research has demonstrated that weight disentanglement is crucial to the success of task arithmetic \cite{ortiz2023task}.
To address this, \cite{yadav2024ties} introduce TIES-Merging, a strategy that minimizes redundant parameters and selects signs to reduce interference during model merging.
In addition, the DARE method proposed by \cite{yu2024language} involves the random dropping of incremental parameters and rescaling operations.
By integrating with other model merging algorithms, it enhances the performance of the model after fusion.
According to the research by \cite{tang2023parameter, stoica2024model}, existing merging methods are more challenging to apply to LoRA fine-tuned models compared to FFT models.
Our method builds upon RegMean and introduces improvements that eliminate the need for obtaining the inner product matrix of training data.
It can also be solved using a closed-form solution, supporting the integration of either FFT models or LoRA adapters.

\section{Method}

\subsection{Motivation}
The RegMean method naturally aims to make the output of the merged model close to the outputs of the original models.
It cleverly uses the inner product matrix derived from the training data through a closed-form solution, while avoiding data privacy issues that may arise from directly using the training data.
However, this brings about additional requirements for model release, as open-source models typically do not provide the inner product matrix.
This limitation significantly impacts the applicability of the RegMean method.

Therefore, we propose a different approach: instead of measuring the output discrepancies between the merged model and the original models, why not directly measure the model weights themselves?
However, directly measuring model weights is equivalent to the weighted averaging method, which has limited effectiveness.
We draw inspiration from research on model pruning, where compression is achieved by removing connections with small weights \cite{han2015deep}.
This aligns with our intuition that larger task vectors generally contain more information.
We adopt a holistic, layer-wise metric based on the squared Frobenius norm, which measures the overall magnitude of weight matrices and serves as an effective indicator of model differences.
For this reason, we use the Frobenius norm of the task vector to quantify model differences and raise it to the power of $k$ to serve as a weighting coefficient, with $k$ as a hyperparameter controlling this balance.

\subsection{Merging of FFT Models}
\begin{figure}[!tbp]
  \centering
  \includegraphics[width=1.0\columnwidth]{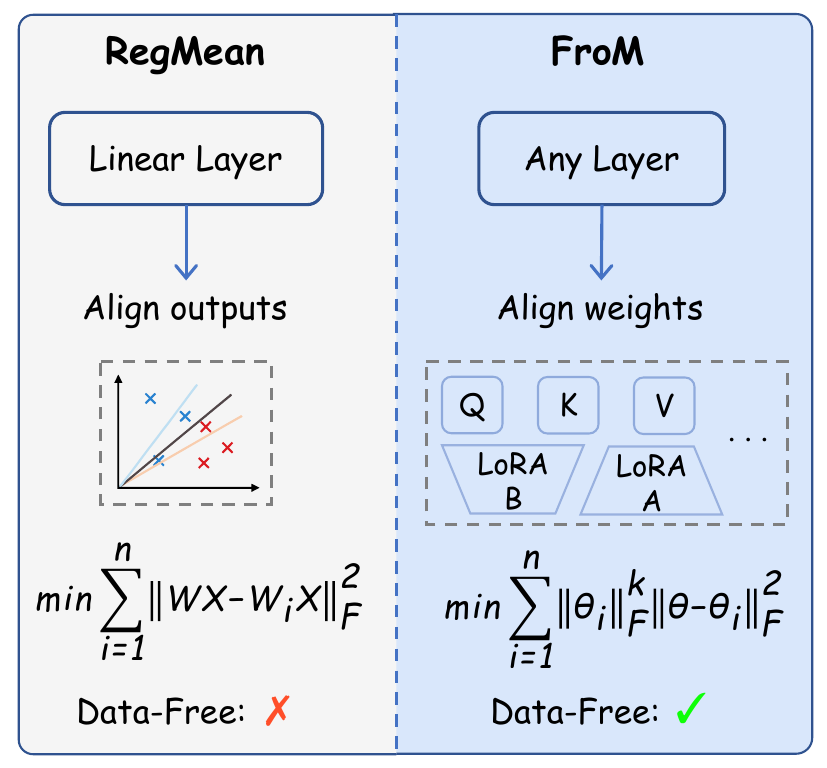}
  \caption{Comparison between the RegMean and FroM Methods, where $X$ represents the input to the linear layer.}
  \label{fig:comparison_regmean}
\end{figure}

We use the objective function in Equation~\ref{eq:fft_loss} to measure the performance difference between the task vectors before and after merging.

\begin{equation}
\mathcal{L}(\boldsymbol{\theta}) = \sum_{i=1}^n \| \boldsymbol{\theta}_i \|_F^k \| \boldsymbol{\theta} - \boldsymbol{\theta}_i \|_F^2
\label{eq:fft_loss}
\end{equation}
where $n$ denotes the number of fine-tuned models, $\boldsymbol{\theta}_i$ represents the task vector of the $i$-th model, $\boldsymbol{\theta}$ is the task vector after merging, and $k$ is a hyperparameter.
When $k = 0$, the objective function of FroM serves as an upper bound for that of RegMean: $\sum_{i=1}^{n} \| \boldsymbol{W}\boldsymbol{X} - \boldsymbol{W}_i \boldsymbol{X} \|_F^2 \leq \sum_{i=1}^{n} \| \boldsymbol{W} - \boldsymbol{W}_i \|_F^2 \cdot \| \boldsymbol{X} \|_F^2 = \sum_{i=1}^{n} \| \boldsymbol{\theta} - \boldsymbol{\theta}_i \|_F^2 \cdot \| \boldsymbol{X} \|_F^2$, where $\| \boldsymbol{X} \|_F^2$ acts as a scaling factor.
Moreover, by setting $k>0$, FroM enables more flexible control in determining an appropriate trade-off point.
The comparison between the RegMean and FroM Methods is illustrated in Figure~\ref{fig:comparison_regmean}.

Since the above function is convex, a closed-form solution (i.e., the point where the gradient is zero) can be derived.
The objective function can be restructured, resulting in:
\begin{align}
\mathcal{L}(\boldsymbol{\theta}) &= \sum_{i=1}^n \| \boldsymbol{\theta}_i \|_F^k \operatorname{Tr} \left( (\boldsymbol{\theta} - \boldsymbol{\theta}_i)^\top (\boldsymbol{\theta} - \boldsymbol{\theta}_i) \right) \\
&= \sum_{i=1}^n \| \boldsymbol{\theta}_i \|_F^k \operatorname{Tr} \left( \boldsymbol{\theta}^\top \boldsymbol{\theta} - 2 \boldsymbol{\theta}^\top \boldsymbol{\theta}_i + \boldsymbol{\theta}_i^\top \boldsymbol{\theta}_i \right)
\end{align}
The last term in the parentheses is a constant; therefore, the derivative with respect to \( \boldsymbol{\theta} \) is:
\begin{align}
\frac{\partial \mathcal{L}(\boldsymbol{\theta})}{\partial \boldsymbol{\theta}} = 2 \sum_{i=1}^n \| \boldsymbol{\theta}_i \|_F^k \boldsymbol{\theta} - 2 \sum_{i=1}^n \| \boldsymbol{\theta}_i \|_F^k \boldsymbol{\theta}_i
\end{align}
By setting $\frac{\partial \mathcal{L}(\boldsymbol{\theta})}{\partial \boldsymbol{\theta}} = 0$, the closed-form solution is derived, as shown in Equation~\ref{eq:fft_sol}.
In fact, we observe that as $k \to 0$, the final result corresponds to the simple averaging method, whereas as $k \to +\infty$, the result converges to the method that selects the weight with the largest norm.
Therefore, this approach can be viewed as a balance between these two methods.

\begin{equation}
\boldsymbol{\theta}^* = \frac{\sum_{i=1}^{n} \| \boldsymbol{\theta}_i \|_F^k \boldsymbol{\theta}_i}{\sum_{i=1}^{n} \| \boldsymbol{\theta}_i \|_F^k}
\label{eq:fft_sol}
\end{equation}

\subsection{Merging of LoRA}

The task arithmetic-based approach poses challenges in integrating LoRA \cite{tang2023parameter, stoica2024model}, making it difficult both to merge and to control the rank of the merged model.
To address this issue, we modify the objective function based on Equation~\ref{eq:fft_loss}, incorporating LoRA-type structures while measuring weight differences using the Frobenius norm.
The revised objective function is given in Equation~\ref{eq:lora_loss}, where $\theta_i$ refers to the task vector in its original form before being flattened into a one-dimensional vector.
When merging parameters layer-wise across different models, we employ LoRA-type structures, specifically utilizing the $\mathbf{A}$ and $\mathbf{B}$ matrices as components optimized in a manner similar to the LoRA mechanism.
Moreover, by explicitly specifying the rank of $\mathbf{A}$ and $\mathbf{B}$ (i.e., the row count of $\mathbf{A}$ and the column count of $\mathbf{B}$), we enable dynamic rank adjustment.
This approach not only increases flexibility in model merging but also facilitates a certain degree of compression during model storage.

\begin{equation}
\mathcal{L}(\mathbf{A}, \mathbf{B}) = \sum_{i=1}^n \| \boldsymbol{\theta}_i \|_F^k \| \mathbf{B} \mathbf{A} - \boldsymbol{\theta}_i \|_F^2
\label{eq:lora_loss}
\end{equation}

Since Equation~\ref{eq:lora_loss} is a non-convex function, it may have multiple local minima.
Therefore, inspired by the Alternating Direction Method of Multipliers in convex optimization theory and analogous to LoRM \cite{salami2024closed}, we propose an alternating optimization algorithm.
Specifically, by initializing $\mathbf{A}$ from a normal distribution with mean $0$ and variance $\sigma^2$, and treating either $\mathbf{A}$ or $\mathbf{B}$ as an invariant, we iteratively update $\mathbf{B}$ and $\mathbf{A}$ using Equation~\ref{eq:lora_solB} and Equation~\ref{eq:lora_solA}, respectively.
In these equations, $\dagger$ denotes the Moore-Penrose inverse (see Appendix~\ref{sec:appendix_a} for the proof).
This procedure for approximating a local minimum of the objective function is shown in Algorithm~\ref{alg:lora_update}.
Due to the potential computational errors in the calculation of the MP inverse, an early stopping mechanism is implemented, which halts the iteration when the loss begins to increase.

\begin{equation}
\mathbf{B} = \left( \sum_{i=1}^n \| \boldsymbol{\theta}_i \|_F^k \boldsymbol{\theta}_i \mathbf{A}^\top \right) \left( \sum_{i=1}^n \| \boldsymbol{\theta}_i \|_F^k \mathbf{A} \mathbf{A}^\top \right)^\dagger
\label{eq:lora_solB}
\end{equation}

\begin{equation}
\mathbf{A} = \left( \sum_{i=1}^n \| \boldsymbol{\theta}_i \|_F^k \mathbf{B}^\top \mathbf{B} \right)^{\dagger} \left( \sum_{i=1}^n \| \boldsymbol{\theta}_i \|_F^k \mathbf{B}^\top \boldsymbol{\theta}_i \right)
\label{eq:lora_solA}
\end{equation}

\subsection{Complexity}

We next analyze the computational complexity of the two merging methods discussed above.
We let $n$ denote the number of models to be merged, and $d$ represent the number of parameters in the FFT models.
Therefore, the complexity of merging FFT models is $O(nd)$.
For the merging of LoRA (i.e., Algorithm~\ref{alg:lora_update}), we simplify the analysis by assuming that LoRA is applied to every layer (in practice, it is typically applied to only a subset of layers).
We let $L$ denote the number of layers in the model and $T$ the number of iterations, assuming that each layer's weight matrix has dimensions $(d_1, d_2)$, with the corresponding LoRA matrices $B$ and $A$ are of sizes $(d_1, r)$ and $(r,d_2)$, respectively.
The first term in Equation~\ref{eq:lora_solB} involves matrix multiplication with complexity $O(d_2^2 r)$.
The second term involves both matrix multiplication and matrix inversion, with respective complexities $O(d_2 r^2)$ and $O(r^3)$.
Thus, the overall complexity becomes: $O(d_2^2 r + d_2 r^2 + r^3) = O(d_2^2 r)$, since $r \ll min(d_1, d_2)$.
Similarly, the complexity of Equation~\ref{eq:lora_solA} is $O(d_1^2 r)$.
Therefore, the overall complexity of the algorithm is: $O(nLTr(d_1^2 + d_2^2))$.

\section{Experimental Setup}

\begin{algorithm}[!tbp]
\caption{Alternating Optimization-Based Model Merging Algorithm Utilizing LoRA-Type Structures}
\label{alg:lora_update}
\begin{algorithmic}[1]
\STATE Initialize matrix $\mathbf{A} \sim \mathcal{N}(0, \sigma^2)$
\STATE Initialize early stopping flag $\texttt{early\_stopping}$
\STATE Set iteration count $\texttt{iters}$
\STATE \( \texttt{min\_loss} \leftarrow \infty \) \COMMENT{Initialize minimum loss}
\FOR{$i = 0$ to $\texttt{iters} - 1$}
    \STATE Compute matrix $\mathbf{B}$ using Equation~\ref{eq:lora_solB}
    \STATE Compute matrix $\mathbf{A}$ using Equation~\ref{eq:lora_solA}
    \IF{\texttt{early\_stopping}}
        \STATE Compute \texttt{loss} using Equation~\ref{eq:lora_loss}
        \IF{\texttt{loss} < \texttt{min\_loss}}
            \STATE \( \texttt{min\_loss} \leftarrow \texttt{loss} \)
        \ELSE
            \STATE \texttt{break}
        \ENDIF
    \ENDIF
\ENDFOR
\STATE Save matrices $\mathbf{A}$ and $\mathbf{B}$
\end{algorithmic}
\end{algorithm}

\paragraph{Merging of FFT Models.}

We test two different model architectures, each based on the same foundational model, across three distinct FFT tasks: Instruction-Following, Mathematical Reasoning, and Code-Generating. 
First, we adopt the experimental setup of \cite{yu2024language}, utilizing models based on Llama-2-13B \cite{touvron2023llama}: WizardLM-13B \cite{xu2023wizardlm}, WizardMath-13B \cite{luo2023wizardmath}, and llama-2-13b-code-alpaca\footnote{https://huggingface.co/layoric/llama-2-13b-code-alpaca}.
Second, we utilize models based on Qwen2.5-7B\footnote{https://huggingface.co/Qwen/Qwen2.5-7B} \cite{yang2024qwen2}: Qwen2.5-7B-Instruct\footnote{https://huggingface.co/Qwen/Qwen2.5-7B-Instruct}, Qwen2.5-Math-7B-Instruct\footnote{https://huggingface.co/Qwen/Qwen2.5-Math-7B-Instruct}, and Qwen2.5-Coder-7B-Instruct\footnote{https://huggingface.co/Qwen/Qwen2.5-Coder-7B-Instruct}.
For testing, we employ the Task Arithmetic \cite{ilharco2022editing}, TIES-Merging \cite{yadav2024ties}, and our FroM methods, additionally incorporating the DARE method \cite{yu2024language} in combination with TIES-Merging and FroM for comparative evaluation.
We do not include the RegMean method in our comparison, as we are unable to obtain the inner product matrices of all model training datasets, which highlights a significant practical limitation of RegMean.
For both evaluation settings involving different base models, we set the linear weighting coefficient $\alpha$ to $1.0$ (see Appendix~\ref{sec:appendix_b2} for details), and use $k = 1.0$ in our FroM method.
For the choice of the hyperparameter $k$ in practical applications, refer to Section~\ref{sec:ablation_study}.

Regarding benchmark selection, we follow the setup of \cite{yu2024language}, utilizing AlpacaEval \cite{alpaca_eval} to assess the models' instruction-following abilities, GSM8K \cite{cobbe2021training} and MATH \cite{hendrycks2021measuring} to evaluate their mathematical reasoning skills, and HumanEval \cite{chen2021evaluating} and MBPP \cite{austin2021program} for testing code generation capabilities.
Further details can be found in Appendix~\ref{sec:appendix_b}.

\paragraph{Merging of LoRA.}

We fine-tune a binary classification head on top of the Meta-Llama-3-8B\footnote{https://huggingface.co/meta-llama/Meta-Llama-3-8B} model.
First, we initialize the classifier by training it on the MNLI dataset from the GLUE benchmark \cite{wang2018glue}.
Then, we fine-tune the model using the binary text entailment datasets from the GLUE benchmark, specifically QNLI, RTE, and WNLI, with the LoRA approach.
For each task, we train the model for 10 epochs and select the best checkpoint for evaluation.
We not only test the merging of LoRA across three tasks but also examine the merging of the three optimal checkpoints for the same task.
We apply the Task Arithmetic, RegMean, TIES-Merging, DARE method (which combines the TIES-Merging approach), KnOTS method (which integrates the first three methods) \cite{stoica2024model}, and FroM method (with $k = 0.9$).
The linear weighting coefficient is set to $\alpha = 0.7$, determined via a linear search.
Further details, including the search results and relevant settings, are provided in Appendix~\ref{sec:appendix_b}.

\section{Main Results}

\subsection{Merging of FFT Models}

\renewcommand{\dblfloatpagefraction}{.9}
\begin{table*}[!htbp]
\centering
\small
\begin{tabular}{@{}clcccccc@{}}
\toprule
\multirow{2}{*}{Models} &
  \multirow{2}{*}{\begin{tabular}[c]{@{}l@{}}Merging\\ Methods\end{tabular}} &
  \begin{tabular}[c]{@{}c@{}}Instruction-\\ Following\end{tabular} &
  \multicolumn{2}{c}{\begin{tabular}[c]{@{}c@{}}Mathematical\\ Reasoning\end{tabular}} &
  \multicolumn{2}{c}{Code-Generating} &
  \multirow{2}{*}{Avg.} \\ \cmidrule(lr){3-7}
 &
   &
  \multicolumn{1}{l}{AlpacaEval} &
  GSM8K &
  \multicolumn{1}{l}{MATH} &
  \multicolumn{1}{l}{HumanEval} &
  \multicolumn{1}{l}{MBPP} &
   \\ \midrule
LM &
  / &
  51.09 &
  45.11 &
  5.92 &
  32.93 &
  31.20 &
  33.25 \\
Math &
  / &
  / &
  59.97 &
  11.60 &
  / &
  / &
  35.78 \\
Code &
  / &
  / &
  / &
  / &
  24.39 &
  28.00 &
  26.20 \\ \midrule
\multirow{5}{*}{\begin{tabular}[c]{@{}c@{}}LM\\ \& Math\end{tabular}} &
  Task Arithmetic &
  50.55 &
  41.62 &
  6.56 &
  / &
  / &
  32.91 \\
 &
  TIES-Merging &
  50.42 &
  51.55 &
  7.22 &
  / &
  / &
  36.40 \\
 &
  DARE+TIES-Merging &
  {\ul 53.03} &
  52.69 &
  \textbf{8.32} &
  / &
  / &
  38.01 \\
 &
  \textbf{FroM} &
  50.82 &
  \textbf{57.39} &
  {\ul 8.00} &
  / &
  / &
  \textbf{38.74} \\
 &
  \textbf{DARE+FroM} &
  \textbf{53.65} &
  {\ul 53.75} &
  7.58 &
  / &
  / &
  {\ul 38.33} \\ \midrule
\multirow{5}{*}{\begin{tabular}[c]{@{}c@{}}LM\\ \& Code\end{tabular}} &
  Task Arithmetic &
  \textbf{53.23} &
  / &
  / &
  32.93 &
  30.60 &
  38.92 \\
 &
  TIES-Merging &
  46.33 &
  / &
  / &
  0.00 &
  0.00 &
  15.44 \\
 &
  DARE+TIES-Merging &
  47.67 &
  / &
  / &
  0.00 &
  0.00 &
  15.89 \\
 &
  \textbf{FroM} &
  {\ul 52.33} &
  / &
  / &
  {\ul 34.76} &
  \textbf{33.40} &
  \textbf{40.16} \\
 &
  \textbf{DARE+FroM} &
  50.40 &
  / &
  / &
  \textbf{35.37} &
  {\ul 32.60} &
  {\ul 39.45} \\ \midrule
\multirow{5}{*}{\begin{tabular}[c]{@{}c@{}}Math\\ \& Code\end{tabular}} &
  Task Arithmetic &
  / &
  \textbf{58.45} &
  \textbf{12.32} &
  5.49 &
  7.00 &
  20.82 \\
 &
  TIES-Merging &
  / &
  58.00 &
  12.28 &
  \textbf{9.15} &
  \textbf{19.00} &
  \textbf{24.61} \\
 &
  DARE+TIES-Merging &
  / &
  57.32 &
  12.10 &
  6.10 &
  {\ul 18.40} &
  23.48 \\
 &
  \textbf{FroM} &
  / &
  {\ul 58.15} &
  11.98 &
  4.88 &
  17.40 &
  23.10 \\
 &
  \textbf{DARE+FroM} &
  / &
  58.07 &
  {\ul 12.30} &
  {\ul 8.54} &
  16.80 &
  {\ul 23.93} \\ \midrule
\multirow{5}{*}{\begin{tabular}[c]{@{}c@{}}LM\\ \& Math\\ \& Code\end{tabular}} &
  Task Arithmetic &
  \textbf{52.84} &
  42.00 &
  6.20 &
  15.85 &
  18.20 &
  27.02 \\
 &
  TIES-Merging &
  46.24 &
  47.08 &
  5.32 &
  0.00 &
  0.00 &
  19.73 \\
 &
  DARE+TIES-Merging &
  48.30 &
  {\ul 55.50} &
  \textbf{9.18} &
  25.00 &
  27.00 &
  33.00 \\
 &
  \textbf{FroM} &
  {\ul 49.65} &
  \textbf{56.48} &
  {\ul 8.20} &
  {\ul 25.61} &
  {\ul 31.40} &
  {\ul 34.27} \\
 &
  \textbf{DARE+FroM} &
  49.12 &
  55.34 &
  8.06 &
  \textbf{29.09} &
  \textbf{31.80} &
  \textbf{34.68} \\ \bottomrule
\end{tabular}
  \caption{\label{tbl:wizard_results}
    Merging results of WizardLM-13B, WizardMath-13B, and llama-2-13b-code-alpaca models based on Llama-2-13B.
    Bold and underlined text indicate the optimal and suboptimal results, respectively.
  }
\end{table*}
\begin{table*}[!htbp]
\centering
\small
\begin{tabular}{@{}clcccccc@{}}
\toprule
\multirow{2}{*}{Models} &
  \multirow{2}{*}{\begin{tabular}[c]{@{}l@{}}Merging\\ Methods\end{tabular}} &
  \begin{tabular}[c]{@{}c@{}}Instruction-\\ Following\end{tabular} &
  \multicolumn{2}{c}{\begin{tabular}[c]{@{}c@{}}Mathematical\\ Reasoning\end{tabular}} &
  \multicolumn{2}{c}{Code-Generating} &
  \multirow{2}{*}{Avg.} \\ \cmidrule(lr){3-7}
 &
   &
  \multicolumn{1}{l}{AlpacaEval} &
  GSM8K &
  \multicolumn{1}{l}{MATH} &
  \multicolumn{1}{l}{HumanEval} &
  \multicolumn{1}{l}{MBPP} &
   \\ \midrule
LM &
  / &
  16.24 &
  79.08 &
  14.40 &
  79.88 &
  65.40 &
  51.00 \\
Math &
  / &
  / &
  75.74 &
  1.12 &
  / &
  / &
  38.43 \\
Code &
  / &
  / &
  / &
  / &
  30.49 &
  39.00 &
  34.74 \\ \midrule
\multirow{5}{*}{\begin{tabular}[c]{@{}c@{}}LM\\ \& Math\end{tabular}} &
  Task Arithmetic &
  \textbf{2.25} &
  \textbf{18.95} &
  \textbf{6.04} &
  / &
  / &
  \textbf{9.08} \\
 &
  TIES-Merging &
  0.00 &
  0.68 &
  0.00 &
  / &
  / &
  0.23 \\
 &
  DARE+TIES-Merging &
  0.00 &
  0.15 &
  0.00 &
  / &
  / &
  0.05 \\
 &
  \textbf{FroM} &
  {\ul 1.95} &
  {\ul 17.51} &
  {\ul 4.08} &
  / &
  / &
  {\ul 7.85} \\
 &
  \textbf{DARE+FroM} &
  0.00 &
  0.45 &
  0.00 &
  / &
  / &
  0.15 \\ \midrule
\multirow{5}{*}{\begin{tabular}[c]{@{}c@{}}LM\\ \& Code\end{tabular}} &
  Task Arithmetic &
  \textbf{42.76} &
  / &
  / &
  {\ul 21.34} &
  {\ul 29.60} &
  {\ul 31.23} \\
 &
  TIES-Merging &
  0.00 &
  / &
  / &
  0.00 &
  0.00 &
  0.00 \\
 &
  DARE+TIES-Merging &
  0.00 &
  / &
  / &
  0.00 &
  0.00 &
  0.00 \\
 &
  \textbf{FroM} &
  {\ul 36.98} &
  / &
  / &
  {\ul \textbf{30.49}} &
  \textbf{42.00} &
  \textbf{36.49} \\
 &
  \textbf{DARE+FroM} &
  0.00 &
  / &
  / &
  0.00 &
  0.00 &
  0.00 \\ \midrule
\multirow{5}{*}{\begin{tabular}[c]{@{}c@{}}Math\\ \& Code\end{tabular}} &
  Task Arithmetic &
  / &
  0.38 &
  0.00 &
  0.00 &
  0.00 &
  0.09 \\
 &
  TIES-Merging &
  / &
  {\ul 1.36} &
  0.00 &
  0.00 &
  0.00 &
  {\ul 0.34} \\
 &
  DARE+TIES-Merging &
  / &
  0.38 &
  0.00 &
  0.00 &
  0.00 &
  0.09 \\
 &
  \textbf{FroM} &
  / &
  \textbf{17.36} &
  0.00 &
  \textbf{1.22} &
  \textbf{0.80} &
  \textbf{4.85} \\
 &
  \textbf{DARE+FroM} &
  / &
  0.38 &
  0.00 &
  0.00 &
  0.00 &
  0.09 \\ \midrule
\multirow{5}{*}{\begin{tabular}[c]{@{}c@{}}LM\\ \& Math\\ \& Code\end{tabular}} &
  Task Arithmetic &
  0.00 &
  0.45 &
  0.00 &
  0.00 &
  0.00 &
  0.09 \\
 &
  TIES-Merging &
  0.00 &
  0.08 &
  0.00 &
  0.00 &
  0.00 &
  0.02 \\
 &
  DARE+TIES-Merging &
  0.00 &
  0.23 &
  0.00 &
  0.00 &
  0.00 &
  0.05 \\
 &
  \textbf{FroM} &
  \textbf{1.22} &
  \textbf{20.55} &
  {\ul \textbf{0.84}} &
  \textbf{2.44} &
  \textbf{2.80} &
  \textbf{5.57} \\
 &
  \textbf{DARE+FroM} &
  0.00 &
  {\ul 0.61} &
  0.00 &
  0.00 &
  0.00 &
  {\ul 0.12} \\ \bottomrule
\end{tabular}
  \caption{\label{tbl:qwen_results}
    Merging results of Qwen2.5-7B-Instruct, Qwen2.5-Math-7B-Instruct, and Qwen2.5-Coder-7B-Instruct models based on Qwen2.5-7B.
    Bold and underlined text indicate the optimal and suboptimal results, respectively.
  }
\end{table*}

The merging results of the WizardLM-13B, WizardMath-13B, and llama-2-13b-code-alpaca models based on Llama-2-13b are shown in Table~\ref{tbl:wizard_results}.
As observed, for the merging of the first two models, our FroM method achieves better average performance, outperforming the previous optimal baseline by 0.73\%.
Similarly, when merging WizardLM-13B and llama-2-13b-code-alpaca, FroM also achieves the highest average score.
For the merging of the latter two models, however, the Task Arithmetic and TIES-Merging methods demonstrate better performance.
In the case of merging three models, the Task Arithmetic method performs best on the AlpacaEval benchmark, but shows moderate performance across the other four datasets.
For tasks related to Mathematical Reasoning and Code-Generating, both our FroM method and its integration with DARE exhibit strong performance, with the latter achieving the best average performance, surpassing the previous optimal baseline by 1.68\%.
Therefore, although the FroM method does not consistently outperform the baseline across every benchmark, a careful selection of the hyperparameter $k$ enables a well-balanced trade-off, ultimately leading to superior performance compared to other baseline methods.

For the base model Qwen2.5-7B, we present the results of merging the models Qwen2.5-7B-Instruct, Qwen2.5-Math-7B-Instruct, and Qwen2.5-Coder-7B-Instruct, as shown in Table~\ref{tbl:qwen_results}.
Unexpectedly, the performance of the merged models deteriorates significantly, except when merging Qwen2.5-7B-Instruct and Qwen2.5-Coder-7B-Instruct using the Task Arithmetic or FroM methods, which yield comparatively better results.
When the other two models are integrated with Qwen2.5-Math-7B-Instruct, a substantial decline in performance is observed.
Based on the specific test outputs, we observe that the majority of the model’s responses are either incoherent or irrelevant to the given task.
This suggests a significant degree of task interference between Qwen2.5-Math-7B-Instruct and the other two models.
After extensive fine-tuning on different datasets, the linear mode connectivity \cite{ainsworth2022git, zhou2023going} among these models tends to be disrupted, leading to the failure of linear weighting methods.
This finding indicates that not all fine-tuned models derived from the same base model are equally suitable for model merging.
Furthermore, the merging results using the TIES-Merging and DARE methods are close to zero (with outputs consisting entirely of garbled text), indicating that these methods aggravate task interference.
In contrast, our FroM method demonstrates greater robustness when merging both two and three models.

Since the training data of these FFT models are not fully released, it is impossible to directly compare the FroM and RegMean methods in Table~\ref{tbl:wizard_results} and Table~\ref{tbl:qwen_results}.
That said, we fully agree that including RegMean as a baseline is crucial for a more comprehensive comparison.
To this end, we conduct additional experiments with RegMean, with detailed results presented in Appendix~\ref{sec:appendix_c}.

\subsection{Merging of LoRA}

For the LoRA merging experiments, we train a shared classification head on the MNLI dataset.
The performance of this classification head on other datasets is presented in the first row of Table~\ref{tbl:lora_results}.
The accuracy across different tasks exceeds 50\%, demonstrating a reasonable degree of generalization, considering that for binary classification tasks, the probability of randomly selecting the correct answer is 50\%.
We further fine-tune on the QNLI, RTE, and WNLI datasets, selecting the best-performing checkpoints for subsequent experiments.

The testing results for different LoRA merging methods are summarized in Table~\ref{tbl:lora_results}.
It is evident that the application of the TIES-Merging and DARE methods results in a decline in model accuracy, indicating that these approaches may not be universally effective across all downstream tasks and may perform particularly poorly on simpler ones.
On the other hand, our FroM method continues to demonstrate effectiveness, with its average performance surpassing other baselines.
Notably, for the QNLI task, our method even exceeds the baseline accuracy, highlighting that our approach enhances the generalization capability of the merged model resulting from multi-task learning.

\begin{table}[!htbp]
\centering
\scriptsize
\begin{tabular}{@{}lllll@{}}
\toprule
Merging Methods         & QNLI  & RTE            & WNLI           & Avg.  \\ \midrule
\multicolumn{5}{c}{\textit{\textbf{Baseline}}}                            \\ \midrule
/                       & 53.74 & 59.21          & 54.93          & 55.96 \\ \midrule
\multicolumn{5}{c}{\textit{\textbf{Fine-tuning}}}                         \\ \midrule
QNLI                    & 94.05 & 56.32          & 43.66          & 64.68 \\
RTE                     & 54.75 & 71.49          & 64.79          & 63.68 \\
WNLI                    & 54.26 & 62.82          & 54.69          & 57.26 \\ \midrule
\multicolumn{5}{c}{\textit{\textbf{3-Model Merging}}}                     \\ \midrule
Task Arithmetic         & 90.10 & \textbf{67.51} & 52.11 & 69.91 \\
RegMean                 & 69.84 & 66.34 & 53.52 & 63.23 \\
TIES-Merging            & 65.50 & 65.70 & \textbf{61.97} & 64.39 \\
DARE+TIES-Merging       & 54.26 & 60.65 & 56.34 & 57.08 \\
KnOTS+Task Arithmetic   & 94.44 & 66.43 & 50.70 & 70.52 \\
KnOTS+TIES-Merging      & 93.65 & 64.98 & 52.11 & 70.25 \\
KnOTS+DARE+TIES-Merging & 59.07 & 63.54 & 60.56 & 61.06 \\
\hdashline[4pt/5pt] \noalign{\vskip 2pt}
\textbf{FroM} & \textbf{94.51} & 64.26 & 53.52 & \textbf{70.76} \\ \bottomrule
\end{tabular}
  \caption{\label{tbl:lora_results}
    Test results of three models fine-tuned using LoRA on QNLI, RTE, and WNLI datasets with different merging methods. Bold text indicates the optimal results.
  }
\end{table}

\begin{figure*}[t]
  \centering
  \includegraphics[width=\textwidth]{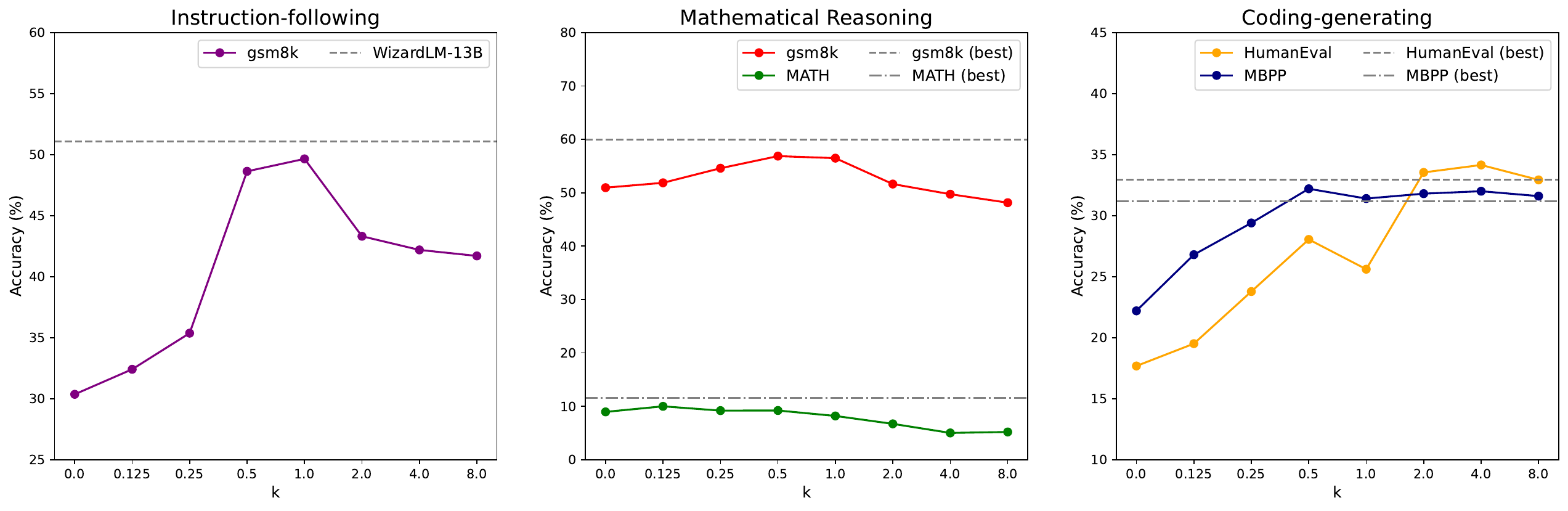}
  \caption{Results of merging WizardLM-13B, WizardMath-13B, and LLaMA-2-13B-Code-Alpaca models with different values of $k$.
  The optimal results before merging are represented by the gray dashed line.}
  \label{fig:alter_norm}
\end{figure*}

\begin{figure}[t]
  \centering
  \includegraphics[width=\columnwidth]{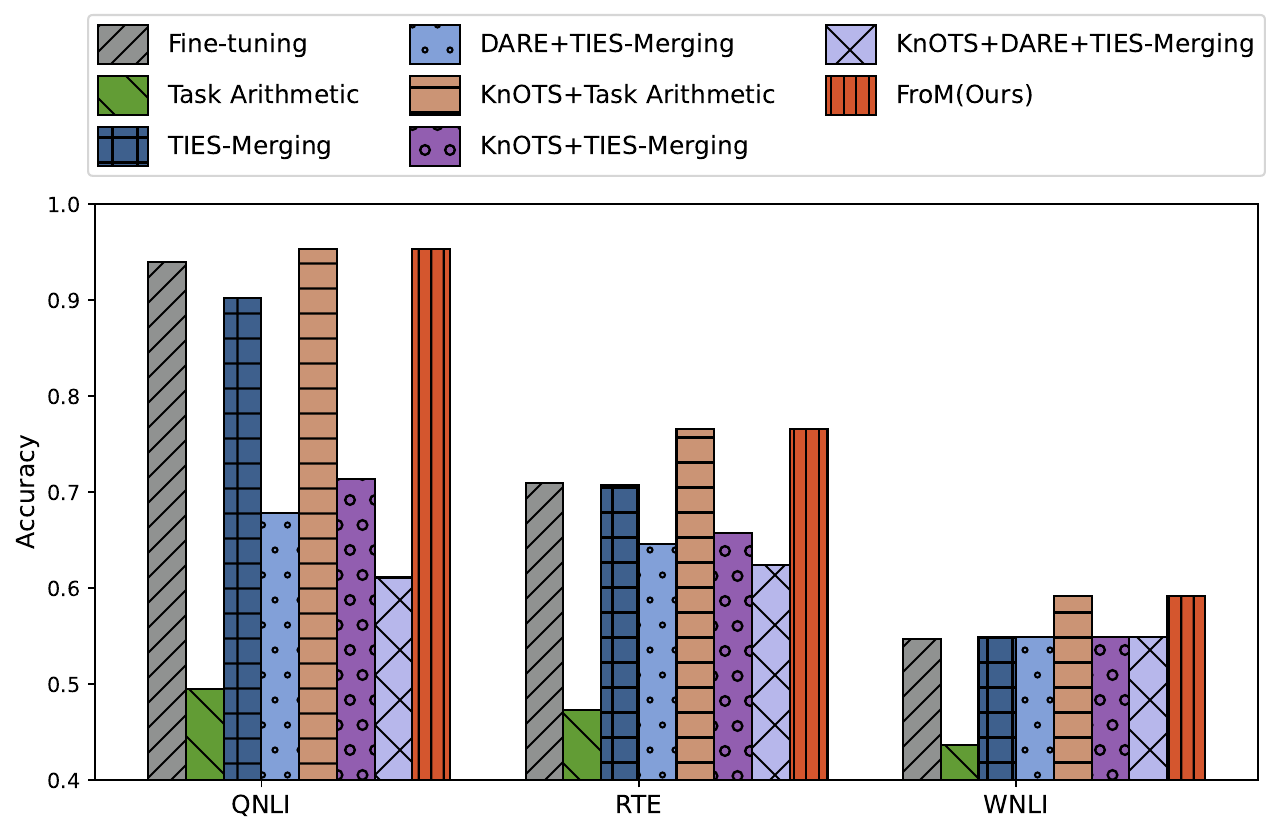}
  \caption{Accuracy comparison after merging the three optimal LoRA checkpoints on QNLI, RTE, and WNLI datasets.}
  \label{fig:lora_merge_checkpoints}
\end{figure}

We also test the merging of the three best checkpoints from the same dataset.
As shown in Figure~\ref{fig:lora_merge_checkpoints}, our FroM method achieves the best performance, surpassing the accuracy of the original fine-tuning across all three datasets.
For comparison, the original KnOTS method performs comparably to our approach, while Task Arithmetic yields significantly inferior results.
The other methods demonstrate relatively moderate performance.
This further demonstrates the effectiveness of our method in merging checkpoints for the same task.

\begin{figure}[!htbp]
  \centering
  \includegraphics[width=\columnwidth]{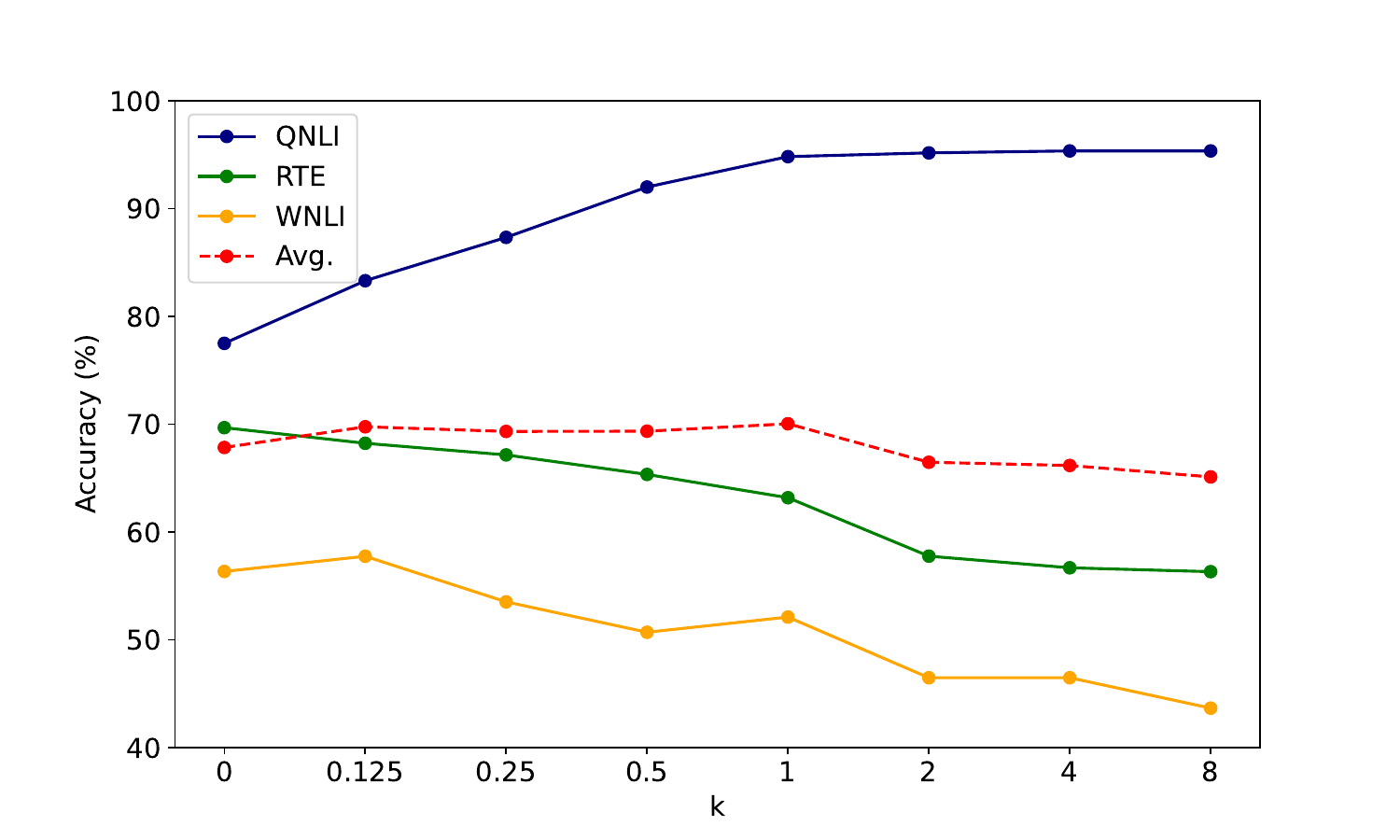}
  \caption{Results of merging the three LoRAs. The average accuracy is depicted by a dashed line.}
  \label{fig:alter_lora_k}
\end{figure}

\subsection{Ablation Study}
\label{sec:ablation_study}

For the WizardLM-13B, WizardMath-13B, and llama-2-13b-code-alpaca models, we test the effect of varying the value of \( k \) in Equation~\ref{eq:fft_loss}. Specifically, we select \( k = 0 \) and integer powers of 2 ranging from \( 2^{-3} \) to \( 2^{3} \), and plot the results under these different conditions.
As shown in Figure~\ref{fig:alter_norm}, the performance is generally poor when \( k \to 0 \) or \( k \to +\infty \).
The best results occur when \( k \) is set to 0.5 or 1.0, with accuracy on all benchmarks closely matching the optimal performance of the fine-tuned models.

We also test the effect of varying the value of $k$ on LoRA fusion, as shown in Figure~\ref{fig:alter_lora_k}.
The optimal average accuracy is achieved when $k$ is set to 1.0.
As the value of $k$ increases, the performance of the fused model improves on the QNLI dataset but decreases on the RTE and WNLI datasets.
This change can be attributed to the larger size of the QNLI dataset, which results in greater influence of its fine-tuned model parameters during merging.

\section{Conclusion}

In this paper, we propose a novel merging approach called FroM, which utilizes the Frobenius norm to quantify weight discrepancies between models without any training data.
This method adjusts adaptively through a hyperparameter, and the merged parameters are derived using a closed-form solution.
For the experimental part, we first test our method on three models based on Llama-2-13B, and the results show that FroM outperforms existing baseline methods.
Next, while baseline methods struggle to effectively integrate three fine-tuned models based on Qwen2.5-7B, our approach outperforms the baseline methods, demonstrating its effectiveness and robustness.
Finally, when merging LoRA adapters, FroM also shows superior performance over other methods.
Overall, the proposed FroM method demonstrates outstanding performance in both FFT and LoRA scenarios, effectively alleviating the task interference issue in model merging and showcasing strong applicability.
We hope this work inspires future research in the model merging field, helps address existing task interference issues, and provides a new perspective for tackling these challenges.

\section*{Limitations}

The limitations of our FroM method can be summarized as follows: (1) the necessity for a more comprehensive theoretical analysis of the algorithm, and (2) the crucial selection of the hyperparameter $k$.
Future research directions with significant potential include providing an interpretable metric for task interference.
By analyzing the upper bound of the model's performance, it may be possible to assess and determine whether the model merging process is feasible.

\section*{Acknowledgments}

Xiaocheng Feng and Bing Qin are the co-corresponding authors of this work. We thank the anonymous reviewers for their insightful comments. This work was supported by the National Natural Science Foundation of China (NSFC) (grant 62276078, U22B2059), the Key R\&D Program of Heilongjiang via grant 2022ZX01A32, and the Fundamental Research Funds for the Central Universities (XNJKKGYDJ2024013).



\bibliography{main}

\begin{thebibliography}{45}
\providecommand{\natexlab}[1]{#1}

\bibitem[{Achiam et~al.(2023)Achiam, Adler, Agarwal, Ahmad, Akkaya, Aleman, Almeida, Altenschmidt, Altman, Anadkat et~al.}]{achiam2023gpt}
Josh Achiam, Steven Adler, Sandhini Agarwal, Lama Ahmad, Ilge Akkaya, Florencia~Leoni Aleman, Diogo Almeida, Janko Altenschmidt, Sam Altman, Shyamal Anadkat, et~al. 2023.
\newblock Gpt-4 technical report.
\newblock \emph{arXiv preprint arXiv:2303.08774}.

\bibitem[{Ainsworth et~al.(2022)Ainsworth, Hayase, and Srinivasa}]{ainsworth2022git}
Samuel~K Ainsworth, Jonathan Hayase, and Siddhartha Srinivasa. 2022.
\newblock Git re-basin: Merging models modulo permutation symmetries.
\newblock \emph{arXiv preprint arXiv:2209.04836}.

\bibitem[{Austin et~al.(2021)Austin, Odena, Nye, Bosma, Michalewski, Dohan, Jiang, Cai, Terry, Le et~al.}]{austin2021program}
Jacob Austin, Augustus Odena, Maxwell Nye, Maarten Bosma, Henryk Michalewski, David Dohan, Ellen Jiang, Carrie Cai, Michael Terry, Quoc Le, et~al. 2021.
\newblock Program synthesis with large language models.
\newblock \emph{arXiv preprint arXiv:2108.07732}.

\bibitem[{Bianchi et~al.(2023)Bianchi, Suzgun, Attanasio, R{\"o}ttger, Jurafsky, Hashimoto, and Zou}]{bianchi2023safety}
Federico Bianchi, Mirac Suzgun, Giuseppe Attanasio, Paul R{\"o}ttger, Dan Jurafsky, Tatsunori Hashimoto, and James Zou. 2023.
\newblock Safety-tuned llamas: Lessons from improving the safety of large language models that follow instructions.
\newblock \emph{arXiv preprint arXiv:2309.07875}.

\bibitem[{Chaudhary(2023)}]{codealpaca}
Sahil Chaudhary. 2023.
\newblock Code alpaca: An instruction-following llama model for code generation.
\newblock \url{https://github.com/sahil280114/codealpaca}.

\bibitem[{Chen et~al.(2021)Chen, Tworek, Jun, Yuan, Pinto, Kaplan, Edwards, Burda, Joseph, Brockman et~al.}]{chen2021evaluating}
Mark Chen, Jerry Tworek, Heewoo Jun, Qiming Yuan, Henrique Ponde De~Oliveira Pinto, Jared Kaplan, Harri Edwards, Yuri Burda, Nicholas Joseph, Greg Brockman, et~al. 2021.
\newblock Evaluating large language models trained on code.
\newblock \emph{arXiv preprint arXiv:2107.03374}.

\bibitem[{Chung et~al.(2024)Chung, Hou, Longpre, Zoph, Tay, Fedus, Li, Wang, Dehghani, Brahma et~al.}]{chung2024scaling}
Hyung~Won Chung, Le~Hou, Shayne Longpre, Barret Zoph, Yi~Tay, William Fedus, Yunxuan Li, Xuezhi Wang, Mostafa Dehghani, Siddhartha Brahma, et~al. 2024.
\newblock Scaling instruction-finetuned language models.
\newblock \emph{Journal of Machine Learning Research}, 25(70):1--53.

\bibitem[{Cobbe et~al.(2021)Cobbe, Kosaraju, Bavarian, Chen, Jun, Kaiser, Plappert, Tworek, Hilton, Nakano et~al.}]{cobbe2021training}
Karl Cobbe, Vineet Kosaraju, Mohammad Bavarian, Mark Chen, Heewoo Jun, Lukasz Kaiser, Matthias Plappert, Jerry Tworek, Jacob Hilton, Reiichiro Nakano, et~al. 2021.
\newblock Training verifiers to solve math word problems.
\newblock \emph{arXiv preprint arXiv:2110.14168}.

\bibitem[{Ding et~al.(2023)Ding, Qin, Yang, Wei, Yang, Su, Hu, Chen, Chan, Chen et~al.}]{ding2023parameter}
Ning Ding, Yujia Qin, Guang Yang, Fuchao Wei, Zonghan Yang, Yusheng Su, Shengding Hu, Yulin Chen, Chi-Min Chan, Weize Chen, et~al. 2023.
\newblock Parameter-efficient fine-tuning of large-scale pre-trained language models.
\newblock \emph{Nature Machine Intelligence}, 5(3):220--235.

\bibitem[{Guo et~al.(2025)Guo, Yang, Zhang, Song, Zhang, Xu, Zhu, Ma, Wang, Bi et~al.}]{guo2025deepseek}
Daya Guo, Dejian Yang, Haowei Zhang, Junxiao Song, Ruoyu Zhang, Runxin Xu, Qihao Zhu, Shirong Ma, Peiyi Wang, Xiao Bi, et~al. 2025.
\newblock Deepseek-r1: Incentivizing reasoning capability in llms via reinforcement learning.
\newblock \emph{arXiv preprint arXiv:2501.12948}.

\bibitem[{Han et~al.(2015)Han, Mao, and Dally}]{han2015deep}
Song Han, Huizi Mao, and William~J Dally. 2015.
\newblock Deep compression: Compressing deep neural networks with pruning, trained quantization and huffman coding.
\newblock \emph{arXiv preprint arXiv:1510.00149}.

\bibitem[{Han et~al.(2021)Han, Zhang, Ding, Gu, Liu, Huo, Qiu, Yao, Zhang, Zhang et~al.}]{han2021pre}
Xu~Han, Zhengyan Zhang, Ning Ding, Yuxian Gu, Xiao Liu, Yuqi Huo, Jiezhong Qiu, Yuan Yao, Ao~Zhang, Liang Zhang, et~al. 2021.
\newblock Pre-trained models: Past, present and future.
\newblock \emph{AI Open}, 2:225--250.

\bibitem[{Han et~al.(2024)Han, Gao, Liu, Zhang, and Zhang}]{han2024parameter}
Zeyu Han, Chao Gao, Jinyang Liu, Jeff Zhang, and Sai~Qian Zhang. 2024.
\newblock Parameter-efficient fine-tuning for large models: A comprehensive survey.
\newblock \emph{arXiv preprint arXiv:2403.14608}.

\bibitem[{Hendrycks et~al.(2021)Hendrycks, Burns, Kadavath, Arora, Basart, Tang, Song, and Steinhardt}]{hendrycks2021measuring}
Dan Hendrycks, Collin Burns, Saurav Kadavath, Akul Arora, Steven Basart, Eric Tang, Dawn Song, and Jacob Steinhardt. 2021.
\newblock Measuring mathematical problem solving with the math dataset.
\newblock \emph{arXiv preprint arXiv:2103.03874}.

\bibitem[{Hu et~al.(2021)Hu, Shen, Wallis, Allen-Zhu, Li, Wang, Wang, and Chen}]{hu2021lora}
Edward~J Hu, Yelong Shen, Phillip Wallis, Zeyuan Allen-Zhu, Yuanzhi Li, Shean Wang, Lu~Wang, and Weizhu Chen. 2021.
\newblock Lora: Low-rank adaptation of large language models.
\newblock \emph{arXiv preprint arXiv:2106.09685}.

\bibitem[{Ilharco et~al.(2022)Ilharco, Ribeiro, Wortsman, Gururangan, Schmidt, Hajishirzi, and Farhadi}]{ilharco2022editing}
Gabriel Ilharco, Marco~Tulio Ribeiro, Mitchell Wortsman, Suchin Gururangan, Ludwig Schmidt, Hannaneh Hajishirzi, and Ali Farhadi. 2022.
\newblock Editing models with task arithmetic.
\newblock \emph{arXiv preprint arXiv:2212.04089}.

\bibitem[{Jin et~al.(2022)Jin, Ren, Preotiuc-Pietro, and Cheng}]{jin2022dataless}
Xisen Jin, Xiang Ren, Daniel Preotiuc-Pietro, and Pengxiang Cheng. 2022.
\newblock Dataless knowledge fusion by merging weights of language models.
\newblock \emph{arXiv preprint arXiv:2212.09849}.

\bibitem[{LI et~al.(2024)LI, Beeching, Tunstall, Lipkin, Soletskyi, Huang, Rasul, Yu, Jiang, Shen, Qin, Dong, Zhou, Fleureau, Lample, and Polu}]{numina_math_datasets}
Jia LI, Edward Beeching, Lewis Tunstall, Ben Lipkin, Roman Soletskyi, Shengyi~Costa Huang, Kashif Rasul, Longhui Yu, Albert Jiang, Ziju Shen, Zihan Qin, Bin Dong, Li~Zhou, Yann Fleureau, Guillaume Lample, and Stanislas Polu. 2024.
\newblock Numinamath.
\newblock \url{[https://huggingface.co/AI-MO/NuminaMath-CoT](https://github.com/project-numina/aimo-progress-prize/blob/main/report/numina_dataset.pdf)}.

\bibitem[{Li et~al.(2023)Li, Zhang, Dubois, Taori, Gulrajani, Guestrin, Liang, and Hashimoto}]{alpaca_eval}
Xuechen Li, Tianyi Zhang, Yann Dubois, Rohan Taori, Ishaan Gulrajani, Carlos Guestrin, Percy Liang, and Tatsunori~B. Hashimoto. 2023.
\newblock Alpacaeval: An automatic evaluator of instruction-following models.
\newblock \url{https://github.com/tatsu-lab/alpaca_eval}.

\bibitem[{Lu et~al.(2024)Lu, Pang, Xiao, Zhu, Xia, and Zhang}]{lu2024merge}
Jinliang Lu, Ziliang Pang, Min Xiao, Yaochen Zhu, Rui Xia, and Jiajun Zhang. 2024.
\newblock Merge, ensemble, and cooperate! a survey on collaborative strategies in the era of large language models.
\newblock \emph{arXiv preprint arXiv:2407.06089}.

\bibitem[{Luo et~al.(2023)Luo, Sun, Xu, Zhao, Lou, Tao, Geng, Lin, Chen, and Zhang}]{luo2023wizardmath}
Haipeng Luo, Qingfeng Sun, Can Xu, Pu~Zhao, Jianguang Lou, Chongyang Tao, Xiubo Geng, Qingwei Lin, Shifeng Chen, and Dongmei Zhang. 2023.
\newblock Wizardmath: Empowering mathematical reasoning for large language models via reinforced evol-instruct.
\newblock \emph{arXiv preprint arXiv:2308.09583}.

\bibitem[{Lv et~al.(2023)Lv, Yang, Liu, Gao, Guo, and Qiu}]{lv2023full}
Kai Lv, Yuqing Yang, Tengxiao Liu, Qinghui Gao, Qipeng Guo, and Xipeng Qiu. 2023.
\newblock Full parameter fine-tuning for large language models with limited resources.
\newblock \emph{arXiv preprint arXiv:2306.09782}.

\bibitem[{Lyu et~al.(2024)Lyu, Zhao, Gu, Yu, Goyal, and Arora}]{lyu2024keeping}
Kaifeng Lyu, Haoyu Zhao, Xinran Gu, Dingli Yu, Anirudh Goyal, and Sanjeev Arora. 2024.
\newblock Keeping {LLMs} aligned after fine-tuning: The crucial role of prompt templates.
\newblock \emph{arXiv preprint arXiv:2402.18540}.

\bibitem[{Matena and Raffel(2022)}]{matena2022merging}
Michael~S Matena and Colin~A Raffel. 2022.
\newblock Merging models with fisher-weighted averaging.
\newblock \emph{Advances in Neural Information Processing Systems}, 35:17703--17716.

\bibitem[{McMahan et~al.(2017)McMahan, Moore, Ramage, Hampson, and y~Arcas}]{mcmahan2017communication}
Brendan McMahan, Eider Moore, Daniel Ramage, Seth Hampson, and Blaise~Aguera y~Arcas. 2017.
\newblock Communication-efficient learning of deep networks from decentralized data.
\newblock In \emph{Artificial intelligence and statistics}, pages 1273--1282. PMLR.

\bibitem[{Ortiz-Jimenez et~al.(2023)Ortiz-Jimenez, Favero, and Frossard}]{ortiz2023task}
Guillermo Ortiz-Jimenez, Alessandro Favero, and Pascal Frossard. 2023.
\newblock Task arithmetic in the tangent space: Improved editing of pre-trained models.
\newblock \emph{Advances in Neural Information Processing Systems}, 36:66727--66754.

\bibitem[{Parthasarathy et~al.(2024)Parthasarathy, Zafar, Khan, and Shahid}]{parthasarathy2024ultimate}
Venkatesh~Balavadhani Parthasarathy, Ahtsham Zafar, Aafaq Khan, and Arsalan Shahid. 2024.
\newblock The ultimate guide to fine-tuning llms from basics to breakthroughs: An exhaustive review of technologies, research, best practices, applied research challenges and opportunities.
\newblock \emph{arXiv preprint arXiv:2408.13296}.

\bibitem[{Salami et~al.(2024)Salami, Buzzega, Mosconi, Bonato, Sabetta, and Calderara}]{salami2024closed}
Riccardo Salami, Pietro Buzzega, Matteo Mosconi, Jacopo Bonato, Luigi Sabetta, and Simone Calderara. 2024.
\newblock Closed-form merging of parameter-efficient modules for federated continual learning.
\newblock \emph{arXiv preprint arXiv:2410.17961}.

\bibitem[{Sanh et~al.(2021)Sanh, Webson, Raffel, Bach, Sutawika, Alyafeai, Chaffin, Stiegler, Scao, Raja et~al.}]{sanh2021multitask}
Victor Sanh, Albert Webson, Colin Raffel, Stephen~H Bach, Lintang Sutawika, Zaid Alyafeai, Antoine Chaffin, Arnaud Stiegler, Teven~Le Scao, Arun Raja, et~al. 2021.
\newblock Multitask prompted training enables zero-shot task generalization.
\newblock \emph{arXiv preprint arXiv:2110.08207}.

\bibitem[{Stoica et~al.(2024)Stoica, Ramesh, Ecsedi, Choshen, and Hoffman}]{stoica2024model}
George Stoica, Pratik Ramesh, Boglarka Ecsedi, Leshem Choshen, and Judy Hoffman. 2024.
\newblock Model merging with svd to tie the knots.
\newblock \emph{arXiv preprint arXiv:2410.19735}.

\bibitem[{Tang et~al.(2023)Tang, Shen, Luo, Zhan, Hu, Du, Chen, and Tao}]{tang2023parameter}
Anke Tang, Li~Shen, Yong Luo, Yibing Zhan, Han Hu, Bo~Du, Yixin Chen, and Dacheng Tao. 2023.
\newblock Parameter efficient multi-task model fusion with partial linearization.
\newblock \emph{arXiv preprint arXiv:2310.04742}.

\bibitem[{Taori et~al.(2023)Taori, Gulrajani, Zhang, Dubois, Li, Guestrin, Liang, and Hashimoto}]{alpaca}
Rohan Taori, Ishaan Gulrajani, Tianyi Zhang, Yann Dubois, Xuechen Li, Carlos Guestrin, Percy Liang, and Tatsunori~B. Hashimoto. 2023.
\newblock Stanford alpaca: An instruction-following llama model.
\newblock \url{https://github.com/tatsu-lab/stanford_alpaca}.

\bibitem[{Touvron et~al.(2023)Touvron, Martin, Stone, Albert, Almahairi, Babaei, Bashlykov, Batra, Bhargava, Bhosale et~al.}]{touvron2023llama}
Hugo Touvron, Louis Martin, Kevin Stone, Peter Albert, Amjad Almahairi, Yasmine Babaei, Nikolay Bashlykov, Soumya Batra, Prajjwal Bhargava, Shruti Bhosale, et~al. 2023.
\newblock Llama 2: Open foundation and fine-tuned chat models.
\newblock \emph{arXiv preprint arXiv:2307.09288}.

\bibitem[{Wang et~al.(2019)Wang, Singh, Michael, Hill, Levy, and Bowman}]{wang2018glue}
Alex Wang, Amanpreet Singh, Julian Michael, Felix Hill, Omer Levy, and Samuel~R. Bowman. 2019.
\newblock \href {https://openreview.net/forum?id=rJ4km2R5t7} {{GLUE}: A multi-task benchmark and analysis platform for natural language understanding}.
\newblock In \emph{International Conference on Learning Representations}.

\bibitem[{Wang et~al.(2020)Wang, Yurochkin, Sun, Papailiopoulos, and Khazaeni}]{wang2020federated}
Hongyi Wang, Mikhail Yurochkin, Yuekai Sun, Dimitris Papailiopoulos, and Yasaman Khazaeni. 2020.
\newblock Federated learning with matched averaging.
\newblock \emph{arXiv preprint arXiv:2002.06440}.

\bibitem[{Wei et~al.(2021)Wei, Bosma, Zhao, Guu, Yu, Lester, Du, Dai, and Le}]{wei2021finetuned}
Jason Wei, Maarten Bosma, Vincent~Y Zhao, Kelvin Guu, Adams~Wei Yu, Brian Lester, Nan Du, Andrew~M Dai, and Quoc~V Le. 2021.
\newblock Finetuned language models are zero-shot learners.
\newblock \emph{arXiv preprint arXiv:2109.01652}.

\bibitem[{Wortsman et~al.(2022)Wortsman, Ilharco, Gadre, Roelofs, Gontijo-Lopes, Morcos, Namkoong, Farhadi, Carmon, Kornblith et~al.}]{wortsman2022model}
Mitchell Wortsman, Gabriel Ilharco, Samir~Ya Gadre, Rebecca Roelofs, Raphael Gontijo-Lopes, Ari~S Morcos, Hongseok Namkoong, Ali Farhadi, Yair Carmon, Simon Kornblith, et~al. 2022.
\newblock Model soups: averaging weights of multiple fine-tuned models improves accuracy without increasing inference time.
\newblock In \emph{International conference on machine learning}, pages 23965--23998. PMLR.

\bibitem[{Xu et~al.(2023{\natexlab{a}})Xu, Sun, Zheng, Geng, Zhao, Feng, Tao, and Jiang}]{xu2023wizardlm}
Can Xu, Qingfeng Sun, Kai Zheng, Xiubo Geng, Pu~Zhao, Jiazhan Feng, Chongyang Tao, and Daxin Jiang. 2023{\natexlab{a}}.
\newblock Wizardlm: Empowering large language models to follow complex instructions.
\newblock \emph{arXiv preprint arXiv:2304.12244}.

\bibitem[{Xu et~al.(2023{\natexlab{b}})Xu, Xie, Qin, Tao, and Wang}]{xu2023parameter}
Lingling Xu, Haoran Xie, Si-Zhao~Joe Qin, Xiaohui Tao, and Fu~Lee Wang. 2023{\natexlab{b}}.
\newblock Parameter-efficient fine-tuning methods for pretrained language models: A critical review and assessment.
\newblock \emph{arXiv preprint arXiv:2312.12148}.

\bibitem[{Yadav et~al.(2024)Yadav, Tam, Choshen, Raffel, and Bansal}]{yadav2024ties}
Prateek Yadav, Derek Tam, Leshem Choshen, Colin~A Raffel, and Mohit Bansal. 2024.
\newblock Ties-merging: Resolving interference when merging models.
\newblock \emph{Advances in Neural Information Processing Systems}, 36.

\bibitem[{Yang et~al.(2024)Yang, Yang, Zhang, Hui, Zheng, Yu, Li, Liu, Huang, Wei et~al.}]{yang2024qwen2}
An~Yang, Baosong Yang, Beichen Zhang, Binyuan Hui, Bo~Zheng, Bowen Yu, Chengyuan Li, Dayiheng Liu, Fei Huang, Haoran Wei, et~al. 2024.
\newblock Qwen2. 5 technical report.
\newblock \emph{arXiv preprint arXiv:2412.15115}.

\bibitem[{Yu et~al.(2024)Yu, Yu, Yu, Huang, and Li}]{yu2024language}
Le~Yu, Bowen Yu, Haiyang Yu, Fei Huang, and Yongbin Li. 2024.
\newblock Language models are super mario: Absorbing abilities from homologous models as a free lunch.
\newblock In \emph{Forty-first International Conference on Machine Learning}.

\bibitem[{Zhao et~al.(2023)Zhao, Zhou, Li, Tang, Wang, Hou, Min, Zhang, Zhang, Dong et~al.}]{zhao2023survey}
Wayne~Xin Zhao, Kun Zhou, Junyi Li, Tianyi Tang, Xiaolei Wang, Yupeng Hou, Yingqian Min, Beichen Zhang, Junjie Zhang, Zican Dong, et~al. 2023.
\newblock A survey of large language models.
\newblock \emph{arXiv preprint arXiv:2303.18223}.

\bibitem[{Zhou et~al.(2023{\natexlab{a}})Zhou, Lu, Mishra, Brahma, Basu, Luan, Zhou, and Hou}]{zhou2023instruction}
Jeffrey Zhou, Tianjian Lu, Swaroop Mishra, Siddhartha Brahma, Sujoy Basu, Yi~Luan, Denny Zhou, and Le~Hou. 2023{\natexlab{a}}.
\newblock Instruction-following evaluation for large language models.
\newblock \emph{arXiv preprint arXiv:2311.07911}.

\bibitem[{Zhou et~al.(2023{\natexlab{b}})Zhou, Yang, Yang, Yan, and Hu}]{zhou2023going}
Zhanpeng Zhou, Yongyi Yang, Xiaojiang Yang, Junchi Yan, and Wei Hu. 2023{\natexlab{b}}.
\newblock Going beyond linear mode connectivity: The layerwise linear feature connectivity.
\newblock \emph{Advances in neural information processing systems}, 36:60853--60877.

\end{thebibliography}

\newpage
\appendix

\onecolumn
\section*{Appendix}

\section{Proof of the expression}
\label{sec:appendix_a}

The objective function given in Equation~\ref{eq:lora_loss} can be reformulated as follows:
\begin{align*}
\mathcal{L}(\mathbf{A}, \mathbf{B}) &= \sum_{i=1}^n \| \boldsymbol{\theta}_i \|_F^k \| \mathbf{B}\mathbf{A} - \boldsymbol{\theta}_i \|_F^2 \\
&= \sum_{i=1}^n \| \boldsymbol{\theta}_i \|_F^k \operatorname{Tr} \left( (\mathbf{B}\mathbf{A} - \boldsymbol{\theta}_i)^\top (\mathbf{B}\mathbf{A} - \boldsymbol{\theta}_i) \right) \\
&= \sum_{i=1}^n \| \boldsymbol{\theta}_i \|_F^k \operatorname{Tr} \left( \mathbf{A}^\top \mathbf{B}^\top \mathbf{B}\mathbf{A} - \boldsymbol{\theta}_i^\top \mathbf{B}\mathbf{A} - \mathbf{A}^\top \mathbf{B}^\top \boldsymbol{\theta}_i + \boldsymbol{\theta}_i^\top \boldsymbol{\theta}_i \right) \\
&= \sum_{i=1}^n \| \boldsymbol{\theta}_i \|_F^k \operatorname{Tr} (\mathbf{A}^\top \mathbf{B}^\top \mathbf{B}\mathbf{A}) - 2 \sum_{i=1}^n \| \boldsymbol{\theta}_i \|_F^k \operatorname{Tr} (\mathbf{A}^\top \mathbf{B}^\top \boldsymbol{\theta}_i) + \sum_{i=1}^n \| \boldsymbol{\theta}_i \|_F^k \operatorname{Tr} \left( \boldsymbol{\theta}_i^\top \boldsymbol{\theta}_i \right)
\end{align*}
Treating matrix \( \mathbf{A} \) as a constant, we differentiate matrix \( \mathbf{B} \) with respect to it, noting that the final term is a constant, which yields:
\begin{align*}
\frac{\partial \mathcal{L}(\mathbf{A}, \mathbf{B})}{\partial \mathbf{B}} = 2 \sum_{i=1}^n \| \boldsymbol{\theta}_i \|_F^k \mathbf{B}\mathbf{A}\mathbf{A}^\top - 2 \sum_{i=1}^n \| \boldsymbol{\theta}_i \|_F^k \boldsymbol{\theta}_i \mathbf{A}^\top
\end{align*}
Let $\frac{\partial \mathcal{L}(\mathbf{A}, \mathbf{B})}{\partial \mathbf{B}} = 0$, then:
\begin{align*}
\mathbf{B} = \left( \sum_{i=1}^n \| \boldsymbol{\theta}_i \|_F^k \boldsymbol{\theta}_i \mathbf{A}^\top \right) \left( \sum_{i=1}^n \| \boldsymbol{\theta}_i \|_F^k \mathbf{A}\mathbf{A}^\top \right)^\dagger
\end{align*}
Similarly, treating matrix \( \mathbf{B} \) as a constant, we differentiate with respect to matrix \( \mathbf{A} \), yielding:
\begin{align*}
\frac{\partial \mathcal{L}(\mathbf{A}, \mathbf{B})}{\partial \mathbf{A}} = 2 \sum_{i=1}^n \| \boldsymbol{\theta}_i \|_F^k \mathbf{B}^\top \mathbf{B}\mathbf{A} - 2 \sum_{i=1}^n \| \boldsymbol{\theta}_i \|_F^k \mathbf{B}^\top \boldsymbol{\theta}_i
\end{align*}
Let $\frac{\partial \mathcal{L}(\mathbf{A}, \mathbf{B})}{\partial \mathbf{A}} = 0$, then:
\begin{align*}
\mathbf{A} = \left( \sum_{i=1}^n \| \boldsymbol{\theta}_i \|_F^k \mathbf{B}^\top \mathbf{B} \right)^\dagger \left( \sum_{i=1}^n \| \boldsymbol{\theta}_i \|_F^k \mathbf{B}^\top \boldsymbol{\theta}_i \right)
\end{align*}

\twocolumn
\section{Experimental Details}
\label{sec:appendix_b}

\subsection{Fine-tuning Settings}
\label{sec:appendix_b1}

For the integration of LoRA adapters, the hyperparameters are set as follows: the ranks of matrices A and B are fixed at 16, the batch size is set to 32, and the learning rate is set to $1e-5$.
We first train a classification head using the MNLI dataset.
Since MNLI is a three-class task, we modify it to a binary classification output by combining the \texttt{neutral} and \texttt{contradiction} categories into a single \texttt{not\_entailment} class.
For each task within the QNLI, RTE, and WNLI datasets, we train the models for 10 epochs and select the checkpoint with the highest accuracy for subsequent LoRA adapter integration.

\subsection{Test Details}
\label{sec:appendix_b2}

For the integration of FFT models, we perform a linear search over the linear weighting coefficient $\alpha$ used to merge the three Llama-2-13B-based models.
The results are presented in Table~\ref{tbl:linear_weighting_coefficient}.
Based on these results, we select $\alpha = 1.0$ as the final linear weighting coefficient to achieve a more balanced outcome.
For the Qwen2.5-7B-based models, due to their poor merging performance, we also adopt $\alpha = 1.0$ without extensive tuning.

Regarding the evaluation, we conducted tests using AlpacaEval, GSM8K, MATH, HumanEval, and MBPP datasets.
In the case of AlpacaEval, the model’s performance is evaluated using the gpt-3.5-turbo model.
Additionally, the final success rate metric selected is \texttt{length\_controlled\_winrate}, rather than \texttt{win\_rate}.
When merging models from the Qwen-2.5 series, we minimized the use of line breaks in the prompts. Excessive line breaks negatively impact the model's responses, leading to the generation of repetitive and meaningless text.

When merging LoRA adapters, we also employ a linear search strategy to determine the optimal linear weighting coefficient $\alpha$ and the hyperparameter $k$ used in our FroM method. Based on this search, we select $\alpha = 0.7$ and $k = 0.9$. Detailed results are presented in Table~\ref{tbl:lora_linear_weighting_coefficient} and ~\ref{tbl:lora_from_k}.

\begin{table}[t]
\centering
\tiny
\begin{tabular}{lcccccc}
\toprule
$\alpha$ & AlpacaEval     & GSM8K          & MATH           & HumanEval      & MBPP           & Avg.           \\ \midrule
0.5                          & 49.76          & 61.79          & \textbf{10.32} & 6.71           & 4.60           & 26.64          \\
0.6                          & 53.88          & \textbf{63.53} & 9.60           & 8.54           & 3.00           & 27.71          \\
0.7                          & 55.35          & 62.40          & 7.92           & 10.37          & 7.20           & 28.65          \\
0.8                          & 56.58          & 60.05          & 7.18           & 11.59          & 13.80          & \textbf{29.84} \\
0.9                          & \textbf{57.00} & 55.57          & 5.62           & 4.88           & 17.00          & 28.01          \\
1.0                          & 52.84          & 42.00          & 6.20           & \textbf{15.85} & \textbf{18.20} & 27.02          \\ \bottomrule
\end{tabular}
  \caption{\label{tbl:linear_weighting_coefficient}
    Results of the linear search for the linear weighting coefficient used in merging Llama-2-13B-based models.
  }
\end{table}

\begin{table}[t]
\centering
\small
\begin{tabular}{ccccc}
\toprule
$\alpha$ & QNLI  & RTE   & WNLI  & Avg. \\ \midrule
0.5                                   & 70.18          & \textbf{68.23} & 53.52          & 63.98          \\
0.6                                   & 81.59          & 67.87          & \textbf{57.75} & 69.07          \\
0.7                                   & 90.10          & 67.51          & 52.11          & \textbf{69.91} \\
0.8                                   & 93.78          & 66.06          & 49.30          & 69.71          \\
0.9                                   & \textbf{94.84} & 64.98          & 46.48          & 68.77          \\
1.0                                   & 78.69          & \textbf{68.23} & \textbf{57.75} & 68.22          \\ \bottomrule
\end{tabular}
  \caption{\label{tbl:lora_linear_weighting_coefficient}
    Results of the linear search for the linear weighting coefficient used in merging LoRA adapters.
  }
\end{table}

\begin{table}[!htbp]
\centering
\small
\begin{tabular}{ccccc}
\toprule
$k$ & QNLI           & RTE            & WNLI           & Avg.           \\ \midrule
0.5 & 92.00          & 65.34          & 50.70          & 69.35          \\
0.6 & 92.84          & \textbf{66.06} & 49.30          & 69.40          \\
0.7 & 93.54          & 64.98          & 52.11          & 70.21          \\
0.8 & 94.07          & 64.62          & \textbf{53.52} & 70.74          \\
0.9 & 94.51          & 64.26          & \textbf{53.52} & \textbf{70.76} \\
1.0 & \textbf{94.82} & 63.18          & 52.11          & 70.04          \\ \bottomrule
\end{tabular}
  \caption{\label{tbl:lora_from_k}
    Results of selecting different values of $k$ in LoRA merging for our FroM method.
  }
\end{table}

\section{Additional Comparison with RegMean}
\label{sec:appendix_c}

We evaluate scenarios involving the merging of three and four models, in order to better demonstrate the generalization capability of our FroM method.
Specifically, we fine-tune the Meta-Llama-3-8B base model on four tasks using supervised fine-tuning: Instruction Following, Mathematical Reasoning, Code Generation, and Safety.
The corresponding datasets are Alpaca-cleaned \cite{alpaca}, NuminaMath-CoT \cite{numina_math_datasets}, Code Alpaca \cite{codealpaca}, and Saferpaca \cite{bianchi2023safety}.
For evaluation across these four dimensions, we use IFEval \cite{zhou2023instruction}, GSM8K, HumanEval, and DirectHarm4 \cite{lyu2024keeping}, where the metric is the proportion of safe responses.
We merge three and four models to better evaluate the generalization ability of these methods.
The results are shown in Table~\ref{tbl:comparison_with_regmean}.
Since RegMean leverages a large amount of training data, FroM does not outperform RegMean but still achieves better results than the other methods.

The results of the linear search for the linear weighting coefficient and the hyperparameter used in FroM within the range $[0.5, 1.0]$ are presented in Table~\ref{tbl:comparison_with_regmean_alpha} and Table~\ref{tbl:comparison_with_regmean_k}.
The final choices are $\alpha = 0.6$ and $k = 0.5$.

\begin{table*}[!htbp]
\centering
\small
\begin{tabular}{@{}lccccc@{}}
\toprule
Merging Methods    & IFEval         & GSM8K          & HumanEval      & DirectHarm4    & Avg.           \\ \midrule
\multicolumn{6}{c}{\textit{\textbf{Baseline}}}                                                          \\ \midrule
/                  & 21.34          & 38.36          & 35.37          & 57.75          & 38.21          \\ \midrule
\multicolumn{6}{c}{\textit{\textbf{Fine-tuning}}}                                                       \\ \midrule
/                  & 41.13          & 42.76          & 45.73          & 93.00          & 55.66          \\ \midrule
\multicolumn{6}{c}{\textit{\textbf{3-Model Merging}}}                                                   \\ \midrule
Task Arithmetic    & \textbf{44.48} & 48.29          & 44.51          & /              & 45.76          \\
TIES-Merging       & \textbf{44.48} & 39.42          & 43.90          & /              & 42.60          \\
DARE               & 26.38          & 0.00           & 0.00           & /              & 8.79           \\
DARE+TIES-Merging          & 26.38          & 0.00           & 0.00           & /              & 8.79           \\
RegMean            & 43.65          & \textbf{52.08} & 45.12          & /              & \textbf{46.95} \\
\hdashline[4pt/5pt] \noalign{\vskip 2pt}
\textbf{FroM}      & \textbf{44.48} & 49.89          & \textbf{46.34} & /              & 46.90          \\
\textbf{DARE+FroM} & 37.05          & 33.74          & 2.44           & /              & 24.41          \\ \midrule
\multicolumn{6}{c}{\textit{\textbf{4-Model Merging}}}                                                   \\ \midrule
Task Arithmetic    & 43.17          & 51.55          & 36.59          & 71.25          & 50.64          \\
TIES-Merging       & 43.65          & 44.73          & \textbf{45.73} & 64.50          & 49.65          \\
DARE               & 26.38          & 0.00           & 0.00           & 70.50          & 24.22          \\
DARE+TIES-Merging          & 26.38          & 0.00           & 0.00           & 67.25          & 23.41          \\
RegMean            & 43.17          & \textbf{54.97} & 42.07          & \textbf{97.75} & \textbf{59.49} \\
\hdashline[4pt/5pt] \noalign{\vskip 2pt}
\textbf{FroM}      & \textbf{45.20} & 52.01          & 45.12          & 65.25          & 51.90          \\
\textbf{DARE+FroM} & 21.94          & 32.68          & 0.61           & 72.50          & 31.93          \\ \bottomrule
\end{tabular}
\caption{Merging results of three and four models based on Meta-LLaMA-3-8B, in comparison with RegMean.}
\label{tbl:comparison_with_regmean}
\end{table*}

\begin{table}[t]
\centering
\scriptsize
\begin{tabular}{@{}cccccc@{}}
\toprule
$\alpha$           & IFEval         & GSM8K          & HumanEval      & DirectHarm4    & Avg.           \\ \midrule
\multicolumn{6}{c}{\textit{\textbf{3-Model Merging}}}                                        \\ \midrule
$0.5$       & 43.88          & \textbf{49.96} & 43.90          & /              & \textbf{45.91} \\
$0.6$       & \textbf{44.48} & 48.29          & \textbf{44.51} & /              & 45.76          \\
$0.7$       & 41.25          & 47.38          & 32.93          & /              & 40.52          \\
$0.8$       & 41.61          & 41.77          & 20.73          & /              & 34.70          \\
$0.9$       & 40.41          & 37.83          & 9.15           & /              & 29.13          \\
$1.0$       & 37.17          & 33.36          & 2.44           & /              & 24.32          \\ \midrule
\multicolumn{6}{c}{\textit{\textbf{4-Model Merging}}}                                        \\ \midrule
$0.5$       & \textbf{43.65}          & \textbf{52.08}          & \textbf{42.07}          & 64.25          & 50.51          \\
$0.6$       & 43.17          & 51.55          & 36.59          & 71.25          & \textbf{50.64}          \\
$0.7$       & 39.21          & 49.51          & 23.17          & \textbf{74.50}          & 46.60          \\
$0.8$       & 39.57          & 46.47          & 7.93           & 73.50          & 41.87          \\
$0.9$       & 29.14          & 43.06          & 0.61           & 64.75          & 34.39          \\
$1.0$       & 21.94          & 32.68          & 0.61           & 72.75          & 32.00          \\ \bottomrule
\end{tabular}
\caption{Results of the linear search for the weighting coefficient used for additional comparison with RegMean.}
\label{tbl:comparison_with_regmean_alpha}
\end{table}

\begin{table}[t]
\centering
\scriptsize
\begin{tabular}{@{}cccccc@{}}
\toprule
$k$    & IFEval         & GSM8K          & HumanEval      & DirectHarm4    & Avg.           \\ \midrule
\multicolumn{6}{c}{\textit{\textbf{3-Model Merging}}}                                             \\ \midrule
$0.5$            & \textbf{44.48} & \textbf{49.89} & 46.34 & /              & \textbf{46.90} \\
$0.6$            & \textbf{44.48} & 48.07          & 45.73          & /              & 46.09          \\
$0.7$            & 43.41          & 47.01          & \textbf{47.56} & /              & 45.99          \\
$0.8$            & 43.88          & 48.67          & 45.73          & /              & 46.09          \\
$0.9$            & 43.17          & 47.76          & 45.73          & /              & 45.55          \\
$1.0$            & 43.76          & 46.25          & 43.90          & /              & 44.64          \\ \midrule
\multicolumn{6}{c}{\textit{\textbf{4-Model Merging}}}                                             \\ \midrule
$0.5$            & 45.20 & \textbf{52.01} & 45.12          & 65.25          & \textbf{51.90} \\
$0.6$            & 45.44          & 49.66          & 45.73          & 64.25          & 51.27          \\
$0.7$            & \textbf{46.04} & 48.52          & 46.95          & 64.00          & 51.38          \\
$0.8$            & 44.00          & 46.40          & 46.34          & \textbf{69.50} & 51.56          \\
$0.9$            & 43.41          & 47.99          & \textbf{48.17} & 66.75          & 51.58          \\
$1.0$            & 42.45          & 46.40          & 44.51          & 65.00          & 49.59          \\ \bottomrule
\end{tabular}
\caption{Results of the linear search for $k$ used for additional comparison with RegMean.}
\label{tbl:comparison_with_regmean_k}
\end{table}

\end{document}